\documentclass[journal,twoside,web]{ieeecolor}
\usepackage{tmi}
\usepackage{cite}
\usepackage{amsmath,amssymb,amsfonts}
\usepackage{algorithmic}
\usepackage{graphicx}
\usepackage{textcomp}
\usepackage{epstopdf}
\usepackage{color}
\usepackage{graphicx}
\usepackage{epstopdf}
\usepackage{times}
\usepackage{epsfig}
\usepackage{graphicx}
\usepackage{amsmath}
\usepackage{amssymb}

\usepackage{amsmath}
\usepackage{multirow}
\usepackage{subfigure}
\usepackage{booktabs}
\usepackage{color}
\usepackage{booktabs}
\usepackage{threeparttable}
\usepackage{footnote}
\usepackage{hyperref}

\def\BibTeX{{\rm B\kern-.05em{\sc i\kern-.025em b}\kern-.08em
    T\kern-.1667em\lower.7ex\hbox{E}\kern-.125emX}}
\begin{document}
\title{Multi-modal Graph Learning for Disease Prediction}
\author{Shuai Zheng,
        Zhenfeng Zhu$^*$,
        Zhizhe Liu,
        Zhenyu Guo,
        Yang Liu,
        Yuchen Yang,\\
        and Yao Zhao,~\IEEEmembership{Senior Member,~IEEE}
\thanks{This work was supported in part by Science and Technology Innovation 2030 - "New Generation Artificial Intelligence" Major Project under Grant No.2018AAA0102100, in part by the National Natural Science Foundation of China under Grant No.61976018.}
\thanks{S. Zheng, Z. Zhu, Z. Liu, Z. Guo, Y. Liu, and Y. Zhao are with the Institute of Information Science, Beijing Jiaotong University, Beijing 100044, China, and also with the Beijing Key
	Laboratory of Advanced Information Science and Network Technology,
	Beijing 100044, China (e-mail: {zs1997, zhfzhu, zhzliu, zhyguo, yangliu, yzhao}@bjtu.edu.cn). }
\thanks{Y. Yang is with the Department of Biology, Johns Hopkins University Krieger School of Arts and Sciences, Baltimore, MD, USA (e-mail: {yuchen.yang}@jhu.edu).(\textit{$^*$Corresponding author: Zhenfeng Zhu.)}}
}

\maketitle

\begin{abstract}
	
	
	Benefiting from the powerful expressive capability of graphs, graph-based approaches have been popularly applied to handle multi-modal medical data and achieved impressive performance in various biomedical applications.
	For disease prediction tasks, most existing graph-based methods tend to define the graph manually based on specified modality (e.g., demographic information), and then integrated other modalities to obtain the patient representation by Graph Representation Learning (GRL).
However, constructing an appropriate graph in advance is not a simple matter for these methods. Meanwhile, the complex correlation between modalities is ignored. These factors inevitably yield the inadequacy of providing sufficient information about the patient's condition for a reliable diagnosis.
	To this end , we propose an end-to-end \underline{M}ulti-\underline{m}odal \underline{G}raph \underline{L}earning framework (MMGL) for disease prediction with multi-modality. 
	To effectively exploit the rich information across multi-modality associated with the disease, modality-aware representation learning is proposed to aggregate the features of each modality by leveraging the correlation and complementarity between the modalities. 
	Furthermore, instead of defining the graph manually, the latent graph structure is captured through an effective way of adaptive graph learning. It could be jointly optimized with the prediction model, thus revealing the intrinsic connections among samples. 
	Our model is also applicable to the scenario of inductive learning for those unseen data.
	An extensive group of experiments on two disease prediction tasks demonstrates that the proposed MMGL achieves more favorable performance. The code of MMGL is available at \url{https://github.com/SsGood/MMGL}.
\end{abstract}


\section{Introduction}
\label{sec:introduction}
\IEEEPARstart{T}he proliferation of multimodal biomedical data has advanced the studies of Computer Aided Diagnosis (CADx) in recent years. Compared to single modal data, heterogeneous medical data from different modalities can provide more complementary information to each other about the patient's condition, facilitating a more reliable diagnosis~\cite{nature, Ali-Reza, MKL2}. Therefore, the effective utilization of multimodal data is very critical for reliable clinical disease diagnosis.

A number of approaches based on multi-modal learning have been proposed for disease prediction. 
These methods aim to learn multi-modal shared representation or fuse multi-modal features in various ways, such as multi-kernel learning~\cite{MKL2,MKL4, MKL1}, non-negative matrix factorization~\cite{NMF2, NMF1}, and deep neural networks~\cite{DNN2, DNN3}. However, there are still several common issues with these approaches. For the multi-modal medical data, the shared representation learning can capture the modality-shared information but the modality-specified complementary information is underutilized. Furthermore, the relationships of patients are not taken into account sufficiently, which is also important for the diagnosis of patients~\cite{tong2017multi,popGCN}. As a general description, the graph model provides a versatile manner for integration of multimodal information and relation discovery among patients due to the inherent characteristic of the graph~\cite{survey}. Up to now, graph-based methods, particularly graph convolutional networks (GCNs) \cite{GCN,Graphsage}, have been applied in various biomedical applications and disease prediction field, such as Alzheimer prediction \cite{TADPOLE,popGCN,InceptionGCN,multi-hop}, Autism prediction \cite{ABIDE,deepgcn_abide}, and cancer prognosis prediction~\cite{MGNN}.



\begin{figure*}[t]	
	\centering
	\includegraphics[width=5.7in]{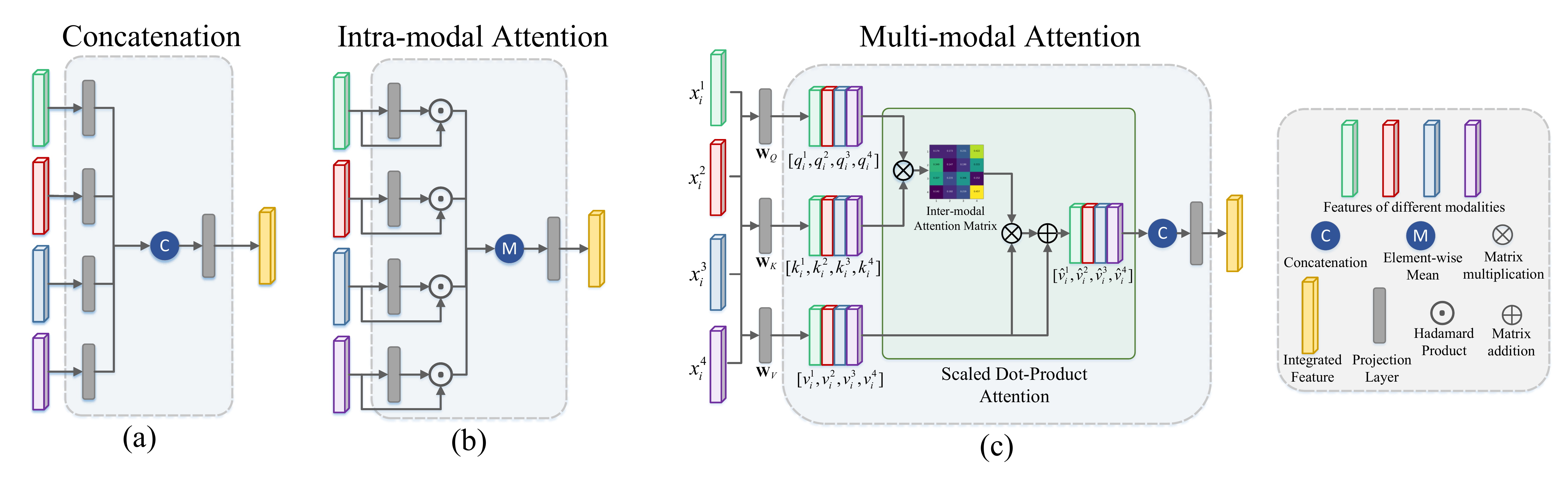}
	
	\caption{Architecture of three types of multi-modal shared representation learning.
		(a) Directly concatenation,
		(b) Intra-modal attention based weighted fusion.
		(c) Our modality-aware representation learning.
		The (a) and (b) fusions have only one interactive operation for different modal features, which proceed at the end of the module. In contrast, our module has more interactive operations through multi-modal attention for features from different modalities. 
	}
	
	\label{fusion}
\end{figure*}

Most existing graph-based methods try to construct the patient relationship graph from existing multi-modal features through pre-defined similarity measures, then apply GCNs to aggregate patient features over local neighborhoods to give the prediction results.
Broadly, these methods can be classified into two categories: single-graph based methods and multi-graph based methods. 
In~\cite{popGCN}, a meta-feature set, such as age and gender, was utilized to calculate patient similarity, so as to construct an adjacency matrix to apply GNNs.
Besides, the influence of graph structure on disease prediction performance was initially discussed by using the same graph construction rule and setting different neighborhood size\cite{InceptionGCN,multi-hop}.
These methods simply combine the imaging and non-imaging modalities through graph construction and GNNs, whereas they fail to effectively mine the intrinsic information of each modality. 
Thus, several recent works have been proposed to construct multiple graphs in parallel, where each graph is built from each modality, then execute the integration of the embeddings learned from different graphs for the prediction.
In summary, as shown in Fig.\ref{fusion}(a), the node embeddings are directly concatenated to form the patient representation\cite{MGNN}. 
In addition, both~\cite{selfGCN} and~\cite{recurrent} adopted attention-based fusion mechanism to integrate the node embeddings as given in Fig.\ref{fusion}(b).

Although the above methods have achieved remarkable performance, three key issues remain to be further considered with respect to the graph-based methods in disease prediction tasks, and even in some other biomedical-related aspects:

\textbf{\emph{(i) Insufficient inter-modal relationship mining.}} Each modality provides different information for the diagnosis of a disease, which explicitly is complementary but also redundant. However, both concatenation~\cite{LGL,MGNN,EV_GCN} and intra-modal attention mechanism~\cite{selfGCN,recurrent} adopted in previous studies are hard to capture the latent \textit{inter-modal correlation}, which may cause the learned representation to be biased towards a single modality. 
In addition, the general multi-modal shared representation learning methods merely focus on capturing the commonalities between modalities, while the dissimilarities between modalities are ignored, possibly resulting in the lack of complementary information. 

\textbf{\emph{(ii) Hand-designing the graph adjacency matrix in a multi-stage framework.}} Both existing single-graph based methods~\cite{yang,popGCN,InceptionGCN,multi-graph2017} and multi-graph based methods \cite{LSTMGCN,MGNN,recurrent} construct the graph through hand-designed similarity measures, which inevitably require careful tuning and are thus difficult to generalize to downstream tasks. Meanwhile, the training of the several parts, such as multimodal representation learning, graph construction, and prediction, are independent of each other in a multi-stage framework. Such practice not only weakens the integrality of the model, but also leads to suboptimal performance in downstream tasks. A better approach is to learn a graph in an adaptive way, which has been studied in GNNs to some extent~\cite{Kipf,adaptive,graph_learning}.  But currently, less focus has been put on the graph structure learning in the biomedical field~\cite{LGL}.


\textbf{\emph{(iii) Hard applicable to inductive learning.}} 
For the approaches based on spectral graph convolution like~\cite{EV_GCN,InceptionGCN,popGCN}, it's hard for them to generalize to unseen samples. Besides, to accommodate the setting of inductive learning, it is also essential but cumbersome for multi-graph based methods \cite{LSTMGCN,MGNN,recurrent} to measure the relationship of unseen samples on each graph. 

To address the issues mentioned above, we concentrate in this paper on graph learning for disease prediction with multi-modality, and the main contributions can be highlighted in the following aspects:
\begin{itemize}
	\item We propose a \underline{M}ulti-\underline{m}odal \underline{G}raph \underline{L}earning model (MMGL) for disease prediction with multi-modality, which is applicable to the scenarios of inductive learning.
	
	\item To characterize a patient with multi-modality, the proposed modality-aware representation learning (MARL) obtains not only the modality-shared representation serving as commonality, but also the modality-specified representation that is patient-sensitive as complementary.
	
	\item To reveal the intrinsic relations among patients, an adaptive graph learning (AGL) is proposed to obtain a latent graph structure to match flexibly for GNN-based downstream tasks. Furthermore, the unified modeling of MARL and AGL can be jointly optimized in an end-to-end way, facilitating more efficient training and inductive testing.
	
	
	
	\item Compared to the state-of-the-art approaches, the comparable even significant improvement on two disease datasets indicates the advantages of our MMGL in terms of disease prediction tasks. Meanwhile, the visualization of contribution score reflected by the obtained dependencies among multi-modality also provides a modal-explainable decision support for doctors in real medical applications and inspiration for disease research.
\end{itemize}
\section{Related Work}

For disease prediction, early multimodal-based studies~\cite{concat3, concat4} usually fused the multi-modal features in the original feature space by the means of direct concatenation. However, the problem is that such heterogeneous data may not be appropriate to be combined directly, leading to the increased risk of dimension curse, significantly. Hence, multi-kernel learning was applied by~\cite{MKL2,MKL4, MKL1} to capture the kernel matrices of each modality, which are combined in a linear or weighted way. Besides,~\cite{DNN2, DNN3} adopted DNN to encode the original feature from each modality, and then perform embedding concatenation to obtain the fused representation. However, little or even no modal interaction have been performed by these methods, i.e., the inter-modal correlations are not explored effectively. Therefore, deep semi-nonnegative matrix factorization has been used to learn modality-shared representation in~\cite{NMF2}. To further mine the relevance between modalities,~\cite{relation-induced} proposed to establish a bi-directional mapping between the original feature and shared embedding to preserve original information. Nevertheless, how to explore the complementary information among multi-modality data remains a key issue for multi-modality based disease diagnosis. In addition to inter-modal relationships, the relationships among patients should also be considered.

Due to the powerful expressive capability of graph for relation modeling, graph-based methods have also been widely used for disease prediction.~\cite{multi-graph2017} proposed to perform graph construction for each modality based on hand-craft kernel and then combined into a unified graph for classification. ~\cite{MGNN} also used GCN to learn patient embedding on constructed multi-graph. In~\cite{popGCN, InceptionGCN}, the non-imaging features were used to construct the population graph while the imaging feature is treated as the individual characteristics to perform neighbor aggregation. Besides,~\cite{selfGCN, recurrent} introduced the attention mechanism on the basis of ~\cite{multi-graph2017} for multi-modal representation fusion. It's worth noting that the hand-crafted graph construction is separated from the prediction module in these methods, leading to cumbersome tuning and poor generalization ability. Meanwhile, they do not further explore the inter-modal relationships except for graph relation construction. Motivated by these observations, our MMGL is proposed to synchronously capture the modality-shared information and the modality-specified information in an end-to-end adaptive graph learning framework.

\section{Methodology}
In this section, we present the methodology of the proposed multi-modal graph learning model.
\subsection{Problem Formulation}
\subsubsection{Notation Definition}



Let $\textbf{X}= \left[x_1, x_2, \cdots, x_N\right]$ denote the raw multi-modal features of $N$ patients and $\textbf{Y}=\left[y_1, y_2, \cdots, y_N\right]$ is the corresponding labels. For the patient $i$ with $M$ modalities, we have $x_i=Concat(x^1_i, x^2_i, \cdots, x^M_i)$, where $x^m_i\in \mathbb{R}^{d_m}$ represents the $m$-th modality of the patient $i$, and $\textbf{X}^m = \left [x^m_1, x^m_2, \cdots, x^m_N \right ]\in \mathbb{R}^{d_m \times N}$ denote the features of the $m$-th modality. 
Treating the patient set as node set $V=\{v_i\}_{i=1}^{N}$ and the connections between each pair of nodes as edge set $E = \{e_{ij}=(v_i,v_j)_{i,j=1}^{N}\}$, a population graph $G = (V, E, \textbf{X})$ can be constructed for disease diagnosis.
In addition, a well defined adjacency matrix $\textbf{A} \in \mathbb{R}^{N \times N}$ is associated with the edge set $E$ and ${A}_{ij}\in \textbf{A}$ represents the edge weight of $e_{ij}$. In addition, for a matrix $\textbf{B}$,  we use $\textup{Vec}(\textbf{B})$ to denote the vectorization  of it by row-wise concatenation.

\subsubsection{Overview of the Framework}

Given the multi-modal medical data $\textbf{X}$, the issue we consider in this paper is to propose a multi-modal graph learning framework for disease prediction considering inter-modal correlations and inter-patient correlations, thus providing appropriate graph learning to support GNNs for disease prediction and the other biomedical tasks. 

As illustrated in Fig.~\ref{Overview}, the overall framework of MMGL consists of three modules, modality-aware representation learning, adaptive graph learning, and GNN-based prediction.
\begin{itemize}
\item[-] \emph{Modality-aware representation learning}. The purpose of the modality-aware representation learning (MARL) is to obtain the modality-aware embeddings $\textbf{H} = concat(\textbf{H}^{sh}, \textbf{H}^{sp})$ from the heterogeneous multi-modal features $\textbf{X}$, which consists of modality-shared information and modality-specified information.
\item[-] \emph{Adaptive graph learning}. Treating the modality-aware embeddings $\textbf{H}$ as a set of graph signals, the adjacency matrix $\textbf{A}$ characterizing the latent population graph is captured through the adaptive graph learning (AGL). 
\item[-] \emph{GNN-based prediction}. Based on the learned modality-aware embeddings $\textbf{H}$ and the adjacency matrix $\textbf{A}$, GNN based model GNN$(\textbf{A},\textbf{H})$ (GCN in our work) can be flexibly applied to provide the prediction $\hat{y}$  of an input  patient $x$ by an inductive way. Along with GNN based prediction, a sub-branch prediction network $f(\cdot)$ is adopted to promote the role of modality-specific embedding and guide the modality-aware representation learning.
\end{itemize}

\subsection{Modality-aware Representation Learning}
\begin{figure*}[t]	
	\centering
	\includegraphics[width=5.2in]{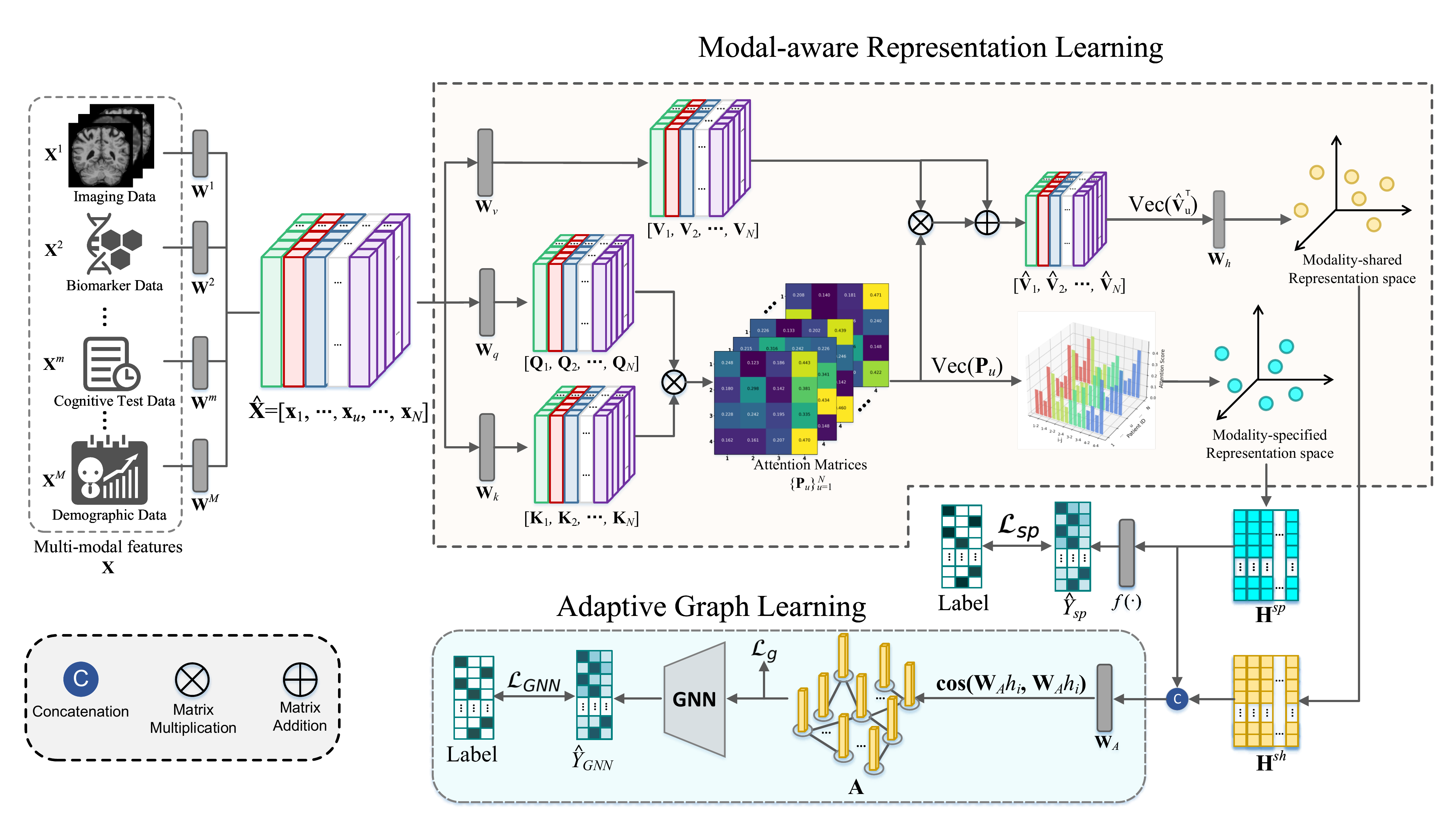}
	\vspace{-0.35cm}
	\caption{The architecture overview of our MMGL. The multi-modal features $\textbf{X}$ is first embedded into the modality-specified representation space and the modality-shared representation space through the modal-aware representation learning. Then an adjacency matrix $\textbf{A}$ for $\textbf{X}$ is learned based on the adaptive graph learning. Finally, we could obtain the prediction results through a GNN based on $A$ and $\textbf{H}$, where $ \textbf{H} = Concat(\textbf{H}^{sh}, \textbf{H}^{sp})$.
	}
	\vspace{-0.5cm}
	\label{Overview}
\end{figure*}
In a real diagnostic scenario, medical experts always need to analyze various multi-modal data of the patient to make a reliable decision, since the single-modal data lack providing enough information for an accurate diagnosis. Similarly, a reliable CADx system also requires to be capable of leveraging the commonality and the complementarity of multi-modal data~\cite{CAD}. Generally, it is thought that commonality among multi-modality implies consistent information about the disease, while complementarity implies other information.
The proposed MMGL model aims to learn a modality-aware representation $\textbf{H}$ from the raw multi-modal features $\textbf{X}$ through a mapping $\mathcal{F} = (\mathcal{F}^{sh}, \mathcal{F}^{sp})$, where $\mathcal{F}^{sh}$ and $\mathcal{F}^{sp}$ are used to capture the common information across multi-modality and modality-dependent specific information for a patient as complementarity, respectively, and we have:  

\begin{align}
\textbf{X} \xrightarrow{\mathcal{F} = (\mathcal{F}^{sh},\hspace{2pt} \mathcal{F}^{sp})} \textbf{H} \doteq (\textbf{H}^{sh}, \textbf{H}^{sp})
\label{function_defining}
\end{align}
where $\textbf{H}^{sh} = \mathcal{F}^{sh}(\textbf{X})$  and $\textbf{H}^{sp} = \mathcal{F}^{sp}(\textbf{X})$ represent  the modality-shared representation and modality-specified representation, respectively. Thus, for a specified patient $u$, the modality-aware representation $h_u$ will consists of $h^{sh}_u$ and $h^{sp}_u$:
\begin{equation}
h_u = Concat(h^{sh}_u, h^{sp}_u)
\label{h_u}
\end{equation}

To this end, we need to capture not only the shared information across multi-modality, but also the complementary information. As an effective way, attention mechanism has been widely used to explore the relationship between elements in different positions in the sequence data or image data. Thus, the multi-modal attention is applied for inter-modal commonality discovery and complementary catching, thus guiding the modal-aware representation learning.   


As illustrated in Fig.\ref{Overview}, to map the heterogeneous raw multi-modal features $\textbf{X}$ to a  $d_f$-dimensional  shared  homogeneous subspace, we first transform each modal features $\{\textbf{X}^m\}_{m=1}^M$ to $\{\hat{\textbf{X}}^m\}_{m=1}^M$ with the identical dimension $d_f$ through the transform matrices $\{\textbf{W}^m\in \mathbb{R}^{{d_m}\times{d_f}}\}_{m=1}^M$. 
Hence, for a patient $u$, the multi-modal feature $x_u$ can be represented as a feature matrix $\textbf{x}_u=[\hat{x}^1_u, \hat{x}^2_u, \cdots, \hat{x}^M_u] \in \mathbb{R}^{d_f \times M}$.
Inspired by the definition of the original attention mechanism for natural language processing in~\cite{transformer, bert}, we treat the modalities as words, therefore the inter-modal attention between two modalities can be obtained by matching the query vector from a given modality with the key vector from the other modality. In practice, considering the problems of space-efficiency and parallelization, the scaled dot product function is chosen as the attention function~\cite{transformer}. Given the feature matrix $\textbf{x}_u$ of the patient $u$, the query matrix $\textbf{Q}_u = \textbf{W}_q \textbf{x}_u =[q^1_u, q^2_u, \cdots, q^M_u]$ can be obtained through the projection matrix $\textbf{W}_q$.
Similarly, the key matrix $\textbf{K}_u$ and the value matrix $\textbf{V}_u$ can be obtained through $\textbf{W}_k$ and $\textbf{W}_v$, respectively. Furthermore, the inter-modal attention matrix $\textbf{P}_u$ is computed as follows:
\begin{equation}
\textbf{P}_{u, ij} = \frac{\mathrm{exp}[(q^i_u)^\top\cdot k^j_u/\tau]}{\sum_{j=1}^{M}\mathrm{exp}[(q^i_u)^\top\cdot k^j_u/\tau]}
\label{attention}
\end{equation}
where $\textbf{P}_{u,ij}$ denotes how much concern the $i$-th modality has for the $j$-th modality of patient $u$, i.e., the dependence relation of them. $\tau$ is the scaling factor to control the hardness of attention, which is set to $\sqrt{d_f}$ like in ~\cite{transformer}. 
On the basis of inter-modal attention matrix $\textbf{P}_u$, the shared cross-modal aggregation for the value vector of each modality could be performed: 
\begin{equation}
\hat{\textbf{V}}_u^{\top} = (\alpha \textbf{I} + \textbf{P}_u)\textbf{V}_u^{\top}
\label{value}
\end{equation}
where $ \textbf{I}$ denotes the identity matrix and $\alpha$ is a hyper-parameter to control the  strength of self-preservation of original modal information. In practice, adding the self-preservation also plays a role of avoiding the gradient vanishing problem in the training process, here we set $\alpha = 1$. Finally, the modality-shared representation $h^{sh}_u \in \mathbb{R}^{d_{h} \times 1}$ is obtained based on the aggregated :
\begin{equation}
h^{sh}_u = \textbf{W}_h \cdot \textup{Vec}(\hat{\textbf{V}}_u^{\top})
\label{h^sh_u}
\end{equation}
where $\textbf{W}_h \in \mathbb{R}^{d_{h} \times d_fM}$ is a projection matrix to perform aggregation of $ \hat{V}_u$. Thus, $\mathcal{F}^{sh}$ in Eq.(\ref{function_defining}) can be defined as:
\begin{equation}
\mathcal{F}^{sh}(\textbf{x}_u) \triangleq \textbf{W}_h \cdot \textup{Vec}((\alpha \textbf{I} + \textbf{P}_u) \textbf{x}_u^{\top} \textbf{W}_v^{\top})
\label{function_shared}
\end{equation}

Although the representation $h^{sh}_u$ is capable of capturing the shared information between modalities of the patient $u$, the inter-modal diversity may be smoothed out in the obtained shared representation. Fortunately, the inter-modal attention matrix $\textbf{P}_u$ not only serves for cross-modal shared aggregation, but also explicitly portrays the patient-sensitive differences between modalities. Specifically, since the $i$-th row of $\textbf{P}_u$ represents the concern of the $i$-th modality to each modality when the $i$-th modality is the primary source of diagnostic basis, the modality-specific information can be reflected by the variance of concern of the $i$-th modality to other modalities. Thus, the attention matrix of each patient could be considered as a representation in the modality-specified embedding space to represent the patient-sensitive information variance of modalities. 
In particular, we define $\mathcal{F}^{sp}$ in Eq.(\ref{function_defining}) as:
\begin{equation}
\mathcal{F}^{sp}(\textbf{x}_u) \triangleq \textup{Vec}(\textbf{P}_u)
\label{function_specific}
\end{equation}
We could obtain the modality-specified embedding $h^{sp}_u = \mathcal{F}^{sp}(\textbf{x}_u) \in \mathbb{R}^{M^2 \times 1}$ to represent the patient-sensitive information as a complement to the shared representation $h^{sh}_u$. Thus, $h_u \in \mathbb{R}^{(d_h + M^2) \times 1}$ could be obtained according to Eq.(\ref{h_u}).

Compared to previous methods~\cite{selfGCN,LGL,MGNN}, the modal-aware representation learning puts concerns on inter-modal correlation and difference through the subject-sensitive attention matrix. As a consequence, it tends to optimally combine the shared and complementary information from different modalities with the consideration of the specificity of each patient. Besides, the proposed multi-modal attention can also be scalable to a multi-head version easily.

\subsection{Adaptive Graph Structure Learning}

Based on the given graph structures, GNNs in~\cite{GCN,GAT,ChebNet} try to learn node representations for downstream tasks through neighborhood aggregation or spectral convolution. However, it's not trivial to obtain an available graph for some specific tasks in the biomedical field. Therefore, the graph learning problem often needs to be considered for applying GNNs in biomedical tasks. 

For graph learning from given feature, it is typically modeled in two forms: \textbf{(i)} learning a joint discrete probability distribution on the edges of the graph~\cite{discrete, graphvae}; \textbf{(ii)} learning a similarity metric of nodes. Since the former is non-differentiable and hard applicable to inductive learning, we take the graph learning problem into consideration from the perspective of similarity metric learning of nodes. Besides,~\cite{graphrnn,graphrann}aim at learning how to generate 
graphs from a set of observed graph, which are not applicable to our task.
Some previous methods have adopted radial basis function (RBF) kernel~\cite{recurrent,multi-hop}, cosine similarity~\cite{EV_GCN}, or threshold-based metric (for discrete feature)~\cite{popGCN,InceptionGCN} as the similarity metric. However, these approaches still require careful manual tuning to construct a meaningful graph structure for downstream GCNs.  Therefore, as illustrated in Fig.~\ref{Overview}, we propose a simple but effective learnable metric function, which could be jointly optimized with the downstream GCNs:

\begin{equation}
\textbf{A}_{ij}= Sim(h_i, h_j) = \cos( \textbf{W}_A h_i,~\textbf{W}_A h_j)
\label{graph}
\end{equation}
where $\textbf{W}_A$ is a learnable weight matrix and $\textbf{A}_{ij}$ is computed as weighted cosine similarity between patient $i$ and $j$. Since there are few uni-directional effects between patients except for epidemics, the learned adjacency matrix $\textbf{A}$ is symmetric that is also in accordance with the expectations of a realistic population graph of patients. 

Commonly, a realistic adjacency matrix is usually non-negative and sparse. Since $\textbf{A}$ is a fully connected graph that is computationally expensive and the element in $\textbf{A}$ is ranging in [-1,1], $\textbf{A}$ is processed to a non-negative sparse graph by graph sparsification with threshold $\theta$. Specifically, we first scale the range of $\textbf{A}$ to [0,1], then set value elements in $\textbf{A}$ that are less than $\theta$ to zero, and finally scale the non-zero elements in $\textbf{A}$ from [$\theta$,1] to [0,1]. That means, we only consider neighbors with link weight larger than $\theta$ for each subject. In practice, we set $\theta=0.5$. In this case, the graph sparsification by the above thresholding is equivalent to applying directly the ReLU function.

The effectiveness of the learned graph $\textbf{A}$ determines the performance of GNN for downstream tasks due to the sensitivity of GNNs to the graph structure. Thus, the constraint on the sparsity, connectivity, and smoothness of the learned graph is also important for adaptive graph learning~\cite{IDGL}. The smoothness constraint $\mathcal{L}_{smh}$ is intended to build links between similar nodes, which means to enforce the smoothness of the graph signals on the learned graph $\textbf{A}$. For the set of graph signals $\{h_1, h_2, \cdots, h_N \}$, the Dirichlet energy is used to measure the smoothness:
\begin{equation}
\mathcal{L}_{smh}(\textbf{A},\textbf{H}) =\frac{1}{2 N^{2}} \sum_{i, j=1}^{N} \textbf{A}_{i j}\left\|h_{i}-h_{j}\right\|^{2}_{2}
\label{smoothness}
\end{equation}
It can be seen that Eq.(\ref{smoothness}) forces the connected nodes in the graph to have similar representations. Essentially, $L_{smh}$ simultaneously serves to control the sparsity of \textbf{A}~\cite{graph_learning}. However, only utilizing $L_{smh}$ may lead to the trivial solution (i.e., $\textbf{A} = 0$). To avoid such situation, an additional regularization term following~\cite{graph_learning} is imposed on $\textbf{A}$:
\begin{equation}
\mathcal{L}_{con}(\textbf{A}) = -\frac{1}{N} \textbf{1}^{\top} \log (\textbf{A} \cdot\textbf{1})
\label{c_and_s}
\end{equation}
where the logarithmic barrier is used to control the connectivity of $\textbf{A}$, and $\textbf{1}\in \mathbb{R}^{N\times1}$ represents an all-one vector. Meanwhile, the Frobenius norm is also applied to $\textbf{A}$ to avoid the excessive sparseness caused by $\mathcal{L}_{smh}$. Specifically, the total graph regularization $\mathcal{L}_{g}$ is defined as:
\begin{equation}
\begin{aligned}
\mathcal{L}_{g}(\textbf{A},\textbf{H}) = \mathcal{L}_{smh}(\textbf{A},\textbf{H}) + \beta\mathcal{L}_{con}(\textbf{A}) + \frac{\gamma}{N^{2}}\|\textbf{A}\|_{F}^{2}
\label{total_graph_regularization}
\end{aligned}
\end{equation}
where $\beta$ and $\gamma$ are two hyper-parameters to balance the regularization terms. Thus, given the modality-aware representations $\textbf{H}$, a graph structure $\textbf{A}$ satisfying sparsity, connectivity, and smoothness could be obtained through minimizing $\mathcal{L}_{g}$:
\begin{equation}
\begin{aligned}
\textbf{A}^* = \mathop{\arg\min}_{\textbf{A}} \mathcal{L}_{g}(\textbf{A},\textbf{H})
\label{total_graph_regularization_summary}
\end{aligned}
\end{equation}
It is worth noting that the patient from the test set is not visible in the training phase due to the inductive learning setting. Nevertheless, with adaptive graph learning, we could implement a dynamic node-relationship measurement through the trained $\textbf{W}_A$ and Eq.(\ref{graph}). Thus, the patients from the test set could be easily added to $\textbf{A}$ during the testing phase.

\vspace{-0.25cm}
\subsection{Model Optimization and Implementation}
\subsubsection{Model optimization} Based on the modality-aware representations $\textbf{H}$ and learned sparse graph $\textbf{A}$, we can use GNNs to output the predicted results $\hat{Y}_{GNN} = GNN(\textbf{A},\textbf{H})$ of the patients. In addition, to stabilize the learning process and promote the modality-specified representation learning, we set an auxiliary classifier $f(\cdot)$ to obtain additional predict results $\hat{Y}_{sp} = f(\textbf{H}^{sp})$ for directly guiding the learning of $H^{sp}$. Without loss of generality, the vanilla GCN structure~\cite{GCN} is adopted as the prediction module, which can also be replaced by other GNNs. For simplicity, the auxiliary classifier $f(\cdot)$ is set to a single-layer MLP.

Unlike~\cite{LGL,EV_GCN} that only optimize the graph structure based on task-aware prediction loss, we use the following joint loss function to guide the optimization of all three modules of MMGL simultaneously:
\begin{equation}
\mathcal{L} = \mathcal{L}_{GNN}(Y, \hat{Y}_{GNN})  + \lambda \mathcal{L}_{g}(\textbf{A},\textbf{H}) + \eta \mathcal{L}_{sp}(Y, \hat{Y}_{sp})
\label{total_loss}
\end{equation}
where $\mathcal{L}_{GNN}$ denotes the task-aware loss based on GNN, $\lambda$ and $\eta$ are hyper-parameters to balance the three loss terms. For the disease prediction task treated as classification problems in our work, $\mathcal{L}_{GNN}$ and $\mathcal{L}_{sp}$ are both set to cross-entropy loss.
\subsubsection{Implementation}
In the training phase, we use a modular iterative training strategy. Specifically, a complete training epoch is performed by jointly training MARL and AGL once, and then jointly training AGL and prediction module once. Through this training strategy, we could simultaneously obtain meaningful patient representations and population graph with high prediction accuracy.

For inductive learning, we adopt the mini-batch training scheme in~\cite{Graphsage} to train the GNN module, instead of training GCN as a whole~\cite{GCN}. Specifically, for a node $u$ from a mini-batch, we first sample its neighbors $\mathcal{N}(u)$ from the graph, and then perform aggregation, which breaks the limitation of aggregation with the entire adjacency matrix in ~\cite{GCN}. 
In the testing phase, we also follow the solution in~\cite{Graphsage} for inductive learning. Specifically, the unseen patient $u$ is first added to the existing population graph by MARL with AGL. Then, similar to the training phase, the neighboring nodes of $u$ are sampled out and aggregated to $u$ for classification. It should be noting that the test patients are not kept in the graph in the testing phase. \textcolor{black}{A two-layer GNN is adopted as the prediction module of MMGL. Thus, we need to sample the 1-hop and 2-hop neighbors of $u$ to perform aggregation. We uniformly use the MultiLayerNeighborSampler function in the DGL library~\cite{DGL} to perform the neighbor sampling in a random manner.} 

The model uses Adam~\cite{Adam} as the optimizer and is implemented on the PyTorch platform. For hyper-parameters tuning,  both $\lambda$ and $\eta$ are set to 1 by experience, and the other hyper-parameters are tuned through hyperopt~\cite{hyperopt}. 

\section{Experimental Results and Analysis}

In this section, we evaluate the performance of MMGL on two challenging biomedical datasets, TADPOLE dataset and ABIDE dataset. We first detail our experimental protocol, and then present the comparison results of MMGL with the state-of-the-art methods.

\subsection{Datasets and Pre-processing}

\begin{table}[t]
\centering
\scriptsize
\caption{Demographic Information of the Baseline Subjects in the TADPOLE Dataset (MMSE: Mini-Mental State Examination)}
\begin{tabular}{@{}lcccc@{}}

\toprule
     & Female / Male & Education & Age & MMSE \\ \midrule
NC   & 106 / 105  & 16.49$\pm$2.58 & 73.49$\pm$5.46 & 29.01$\pm$1.32 \\
sMCI & 121 / 154  & 15.97$\pm$2.73 & 71.96$\pm$7.84 & 27.87$\pm$1.90 \\
pMCI & 20 / 25    & 16.28$\pm$2.70 & 72.66$\pm$6.17 & 23.20$\pm$2.84 \\
AD   & 27 / 45    & 16.26$\pm$2.62 & 74.20$\pm$7.16 & 22.12$\pm$3.39 \\ \bottomrule
\end{tabular}
\label{table::tadpole information}
\vspace{-0.3cm}
\end{table}

\begin{table}[t]
\centering
\caption{Demographic Information of the Baseline Subjects in the ABIDE Dataset}
\begin{tabular}{@{}lccc@{}}
\toprule
         & Female / Male & Age            & \begin{tabular}[c]{@{}c@{}}Eye\_Status\\ (Open / Closed)\end{tabular} \\ \midrule
NC   & 90 / 378      & 16.84$\pm$7.23 & 321 / 147  \\
ASD   & 54 / 349      & 17.07$\pm$7.95 & 288 / 115  \\ \bottomrule
\end{tabular}
\label{table::abide information}
\vspace{-0.4cm}
\end{table}

\subsubsection{TADPOLE} As a subset of the Alzheimer’s Disease Neuroimaging Initiative (ADNI) database, TADPOLE dataset~\cite{TADPOLE} contains features extracted from multi-modality, including MRI, PET, cognitive tests, cerebrospinal fluid (CSF) biomarkers, risk factors, and demographic information. 
For Alzheimer’s Disease prediction, we select 603 subjects with multi-modal features from TADPOLE dataset, divided into 211 Normal Control (NC), 320 Mild Cognitive Impairment (MCI), and 72 Alzheimer’s Disease (AD) patients, respectively. 
Furthermore, part of the MCI subjects would progress to AD over time, which are further labeled as progressive MCI (pMCI). Excluding the pMCI subjects, the rest of the MCI subjects would remain stable, which are labeled as stable MCI (sMCI). Thus, the 320 MCI subjects could be further divided into 275 sMCI and 45 pMCI subjects.

Since each patient may have multiple medical records, only a single sample of the same subject is retained based on subject ID. Then, we remove the longitudinal features and perform feature selection for each modality. Subjects with a feature missing rate greater than 5\% are removed and Mean Imputation is used for missing value filling for remained samples. The demographic information of the selected subjects is shown in Table.~\ref{table::tadpole information}

\subsubsection{ABIDE}
To accelerate understanding of the neural bases of autism, the Autism Brain Imaging Data Exchange \footnote{http://fcon\_1000.projects.nitrc.org/indi/abide/}(ABIDE)~\cite{ABIDE_data} collected over 1000 resting-state functional magnetic resonance imaging (R-fMRI) data with corresponding phenotypic data from 24 different sites. Following the settings in~\cite{popGCN}, we select the same set of 871 subjects for Autism disease prediction. It can be divided into 468 NC and 403 Autism Spectrum Disorder (ASD) subjects. For each subject, a total of four modalities, including demographic information, automated anatomical quality assessment metrics, automated functional quality assessment metrics, and fMRI connection networks, are utilized to evaluate the proposed MMGL model. 

For a fair comparison, we follow the preprocessing step as in~\cite{popGCN}, the Configurable Pipeline for the Analysis of Connectomes (C-PAC)~\cite{ABIDE}. The specific preprocessing steps can be found at this link\footnote{https://github.com/parisots/population-gcn/blob/master/fetch\_data.py}. Table~\ref{table::abide information} shows the demographic information of the subjects.

\renewcommand{\thefootnote}{*}
\begin{table*}[t]
\tiny  
\centering
\caption{ Quantitative comparisons over two benchmark datasets.$(\%)$}
\begin{tabular}{@{}l|cc|cccc|cccc@{}}
\toprule
& \multicolumn{6}{c|}{TADPOLE}                                                 & \multicolumn{4}{c}{ABIDE}             \\ \cmidrule(l){2-11} 
\multicolumn{1}{l|}{Method}                     & \multicolumn{2}{c|}{AD vs. sMCI vs. NC} & \multicolumn{4}{c|}{sMCI vs. pMCI} & \multicolumn{4}{c}{NC vs. ASD} \\ \cmidrule(l){2-11} 
& ACC (\%)             & AUC (\%)               & ACC (\%)          & AUC (\%)          & SEN (\%)          & SPE (\%)          & ACC (\%)         & AUC (\%)         & SEN (\%)        & SPE (\%)\\ \midrule
MLP                              
& 82.28$\pm$4.39       & 83.13$\pm$3.20         & 85.40$\pm$4.47    & 67.69$\pm$8.66    & 87.42$\pm$3.71    & 68.50$\pm$6.30    & 75.22$\pm$8.06   & 79.30$\pm$7.95   & 77.35$\pm$9.00  & 75.24$\pm$10.90    \\
(H)PopGCN~\cite{popGCN}             
& 82.37$\pm$5.10       & 80.71$\pm$4.21         & 79.97$\pm$3.77    & 81.32$\pm$4.11    & 88.32$\pm$2.17    & 85.46$\pm$2.58    & 69.80$\pm$3.35   & 70.32$\pm$3.90   & 73.35$\pm$7.74  & 80.27 $\pm$6.48    \\
(H)InceptionGCN~\cite{InceptionGCN} 
& 77.42$\pm$1.53       & 81.58$\pm$1.31         & 85.81$\pm$3.52    & 85.97$\pm$3.88    & 84.56$\pm$4.14    & 85.00$\pm$3.87    & 72.69$\pm$2.37   & 72.81$\pm$1.94   & 80.29$\pm$5.10  & 74.41 $\pm$6.22     \\
(H)Multi-GCN~\cite{selfGCN}\footnotemark[1]  
& 83.50$\pm$4.91       & 89.34$\pm$5.38         & 86.85$\pm$5.19    & 86.10$\pm$4.75    & 88.23$\pm$5.30    & 82.31$\pm$6.14    & 69.24$\pm$5.90   & 70.04$\pm$4.22   & 70.93$\pm$4.68  & 74.33 $\pm$6.07     \\
(H)LSTMGCN~\cite{LSTMGCN}           
& 83.40$\pm$4.11       & 82.42$\pm$7.97         & 86.53$\pm$6.36    & 85.07$\pm$5.81    & 87.37$\pm$4.58    & 82.85$\pm$3.93    & 74.92$\pm$7.74   & 74.71$\pm$7.92   & 78.57$\pm$11.61 & 78.87 $\pm$7.79     \\
(L)EV-GCN~\cite{EV_GCN}             
& 88.51$\pm$2.34       & 89.97$\pm$2.15         & \underline{88.66$\pm$5.32}    & 86.67$\pm$4.67    & \underline{90.32$\pm$4.10}    & \underline{87.97$\pm$3.96}    & 85.90$\pm$4.47   & 84.72$\pm$4.27   & \underline{88.23$\pm$7.18}  & 79.90$\pm$7.37    \\
(L)LGL~\cite{LGL}\footnotemark[1]      
& \underline{91.37$\pm$2.12}       &\textbf{93.96$\pm$1.45} & 88.30$\pm$4.33    & \underline{90.27$\pm$3.83}    & 90.22$\pm$5.77    & 87.04$\pm$4.61    & \underline{86.40$\pm$1.63}   & \underline{85.88$\pm$1.75}   & 86.31$\pm$4.52  & \underline{88.42$\pm$3.04}    \\ \midrule

MMGL(Ours)                             
& \textbf{92.31$\pm$1.73} & \underline{93.91$\pm$2.10} & \textbf{92.30$\pm$2.93} & \textbf{92.38$\pm$4.11} & \textbf{93.26$\pm$3.61} & \textbf{91.50$\pm$4.22} & \textbf{89.77$\pm$2.72} & \textbf{89.81$\pm$2.56} & \textbf{90.32$\pm$4.21} & \textbf{89.30$\pm$6.04}    \\ \bottomrule
\end{tabular}
\label{table::quantitative comparison}
\vspace{-0.25cm}
\end{table*}

\begin{figure*}[t]
	
	\subfigure[ACC($\%$), TADPOLE]{
		\centering
		\includegraphics[width=1.4in]{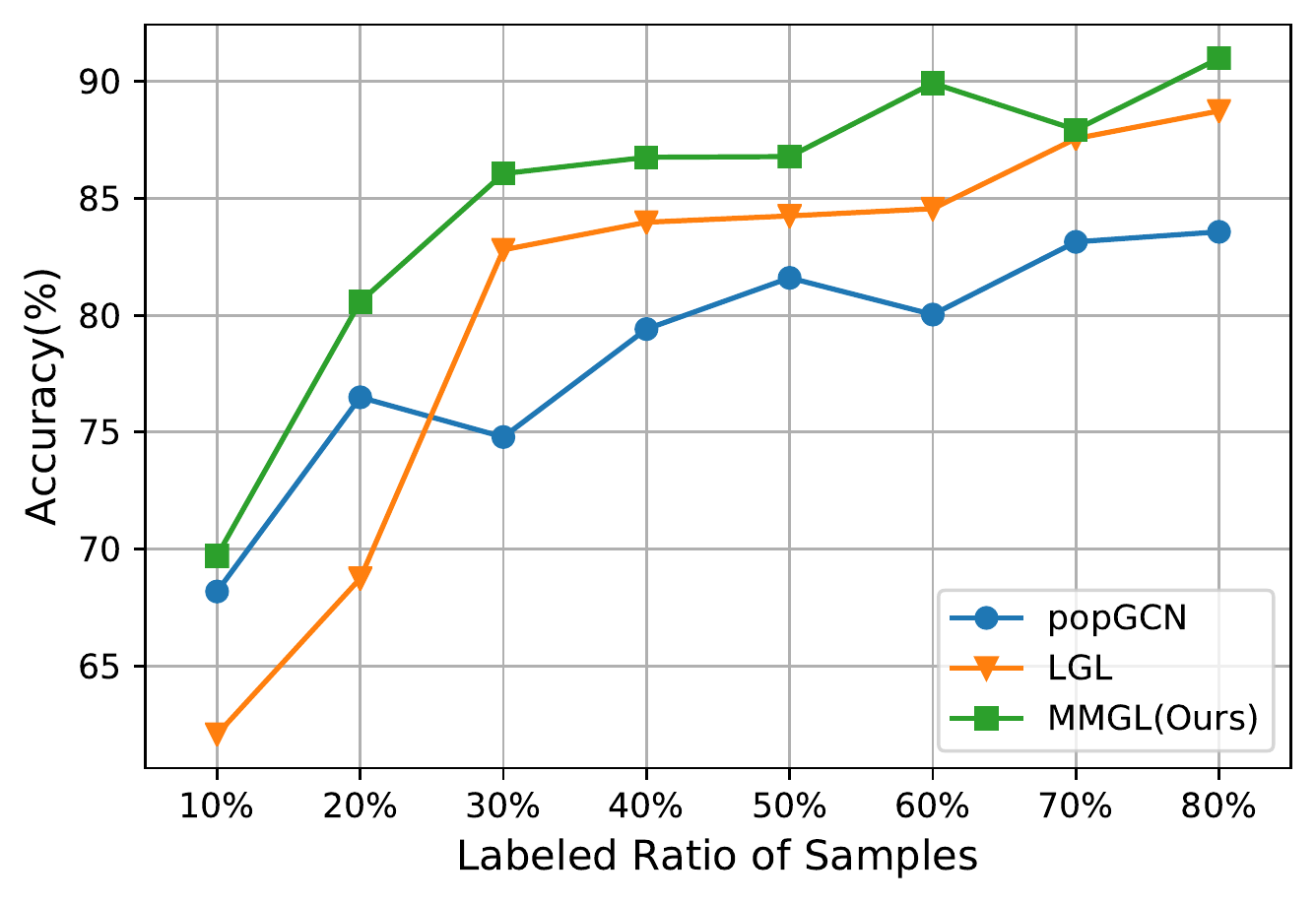}
	}%
	\subfigure[AUC($\%$), TADPOLE]{
		\centering
		\includegraphics[width=1.4in]{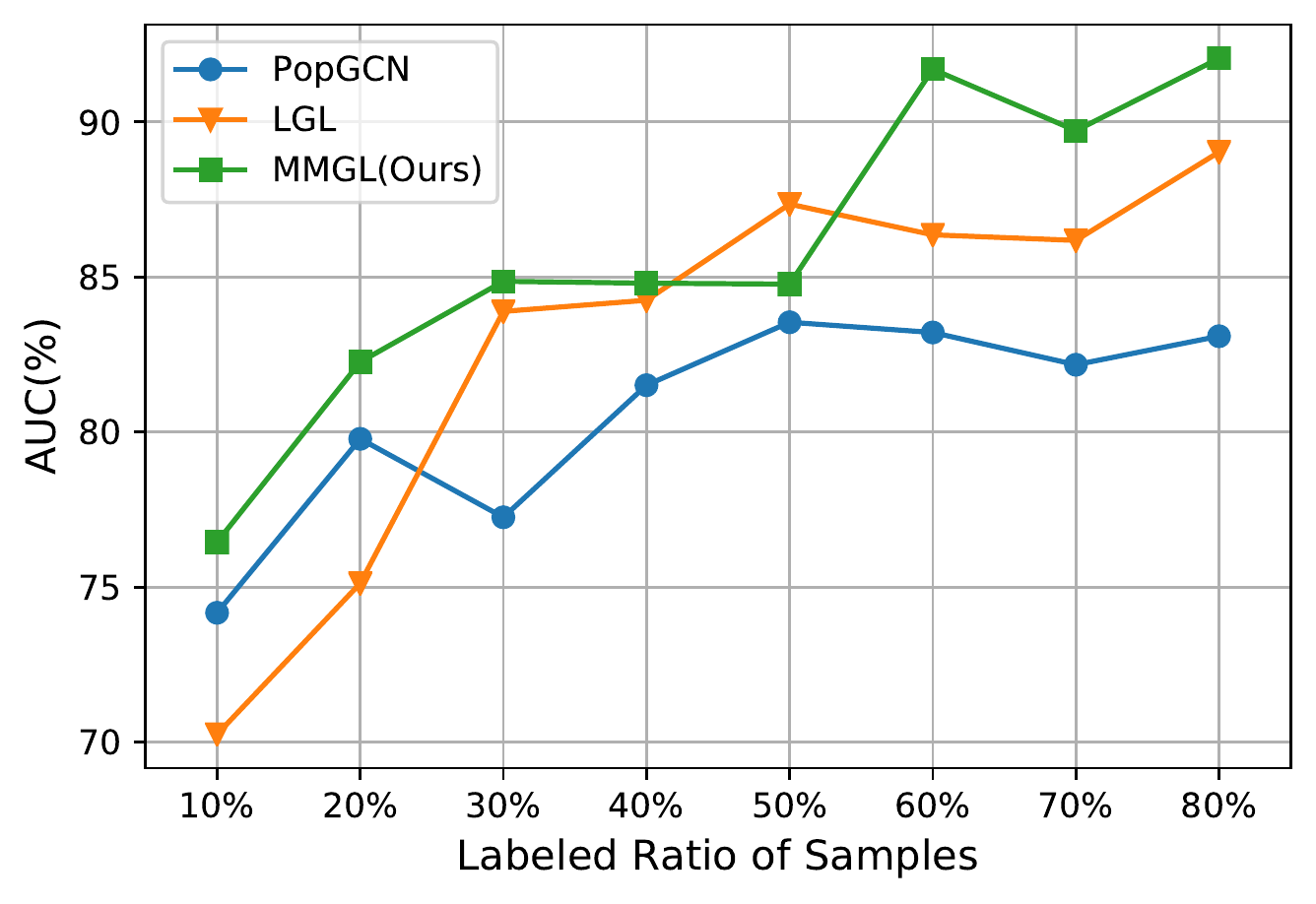}
	}%
	\centering
	\subfigure[ACC($\%$), ABIDE]{
		\centering
		\includegraphics[width=1.4in]{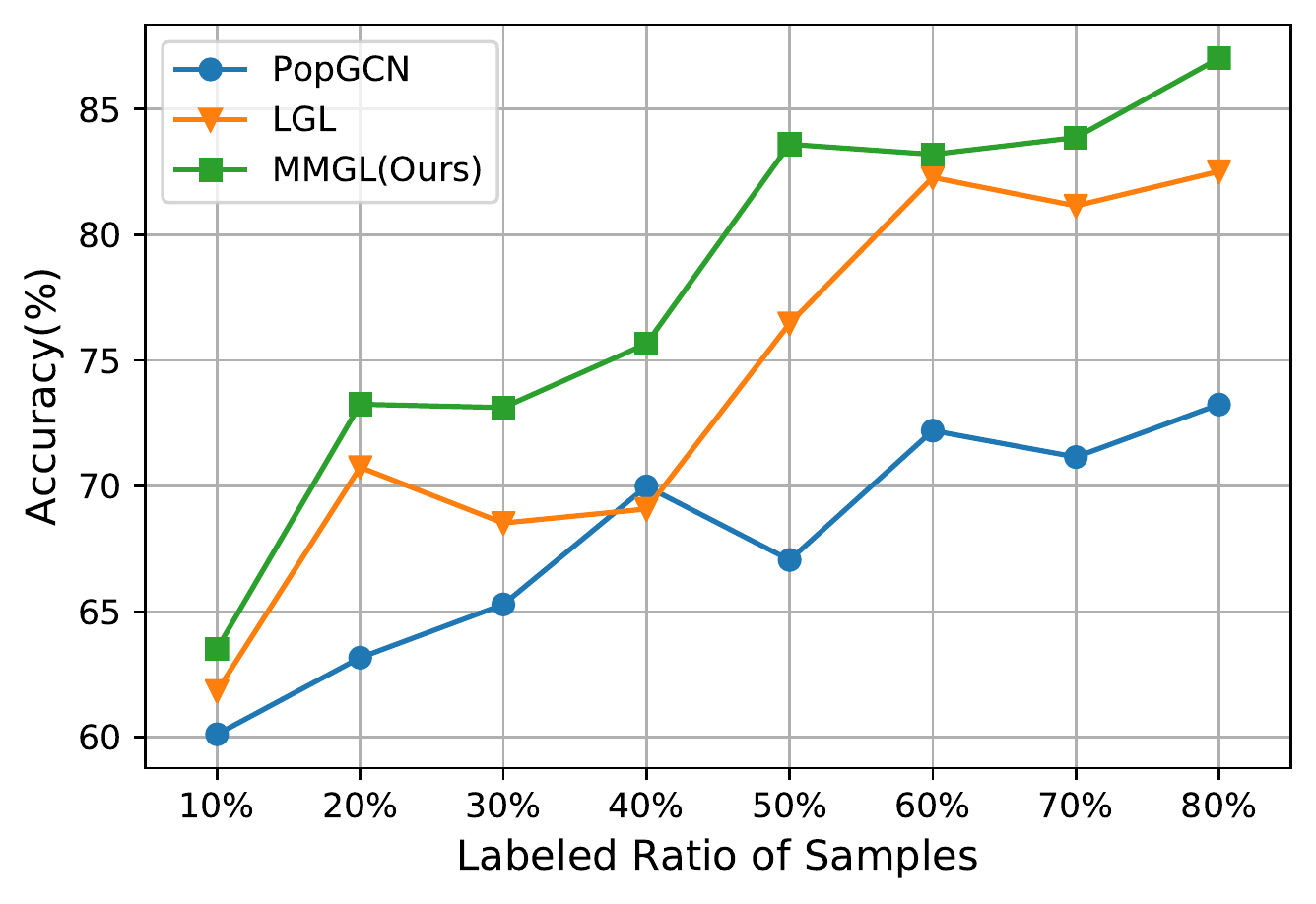}
	}%
	\subfigure[AUC($\%$), ABIDE]{
		\centering
		\includegraphics[width=1.4in]{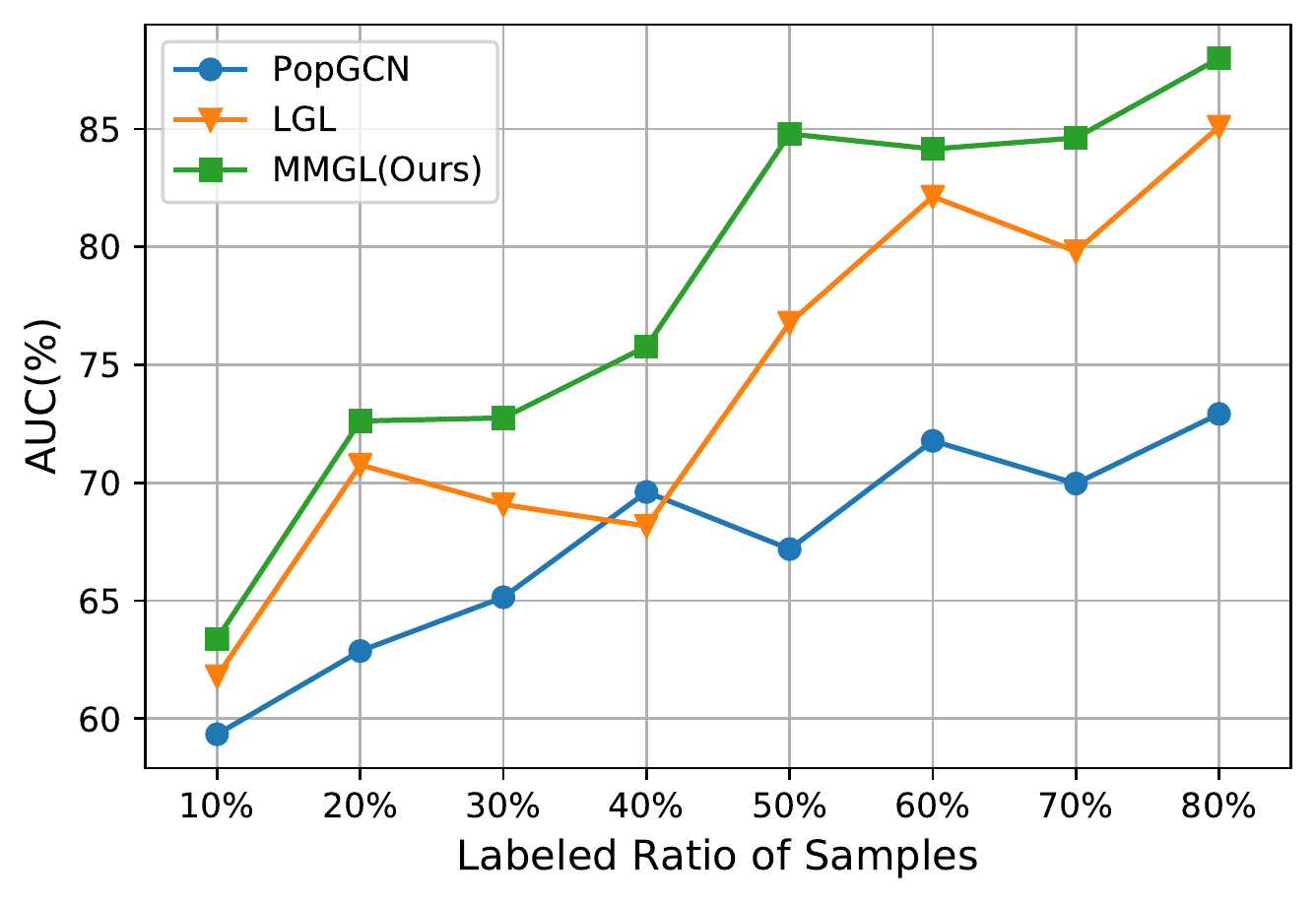}
	}%
	\centering
	\vspace{-0.2cm}
	\caption{Classification performance of the methods with different sizes of training set on transductive learning setting. For TADPOLE dataset, we report the results of the task of NC vs. sMCI vs. AD. For ABIDE dataset, we report the results of the task of NC vs. ASD.}
	\centering
	\label{fig::multi-classificaion}
	\vspace{-0.3cm}
\end{figure*}
\vspace{-0.25cm}
\subsection{Performance Comparisons}
For disease diagnosis on TADPOLE dataset, we evaluate the performance of MMGL with three-classification task, i.e., NC vs. sMCI vs. AD. In addition, we also evaluate MMGL with the sMCI vs. pMCI classification task, which is important for AD early diagnosis. For ABIDE dataset, we evaluate MMGL with two-classification task, NC vs. ASD. 
\subsubsection{Baselines} To evaluate the performance of MMGL, we choose to compare with several baselines, especially those that have achieved the state-of-the-art results in disease prediction tasks recently. 
\begin{itemize}
\item{PopGCN~\cite{popGCN}:} PopGCN uses the demographic information to manually build a population graph, and subsequently aggregates the imaging feature of subjects using GCN for classification.
\item{InceptionGCN~\cite{InceptionGCN}:} On the basis of~\cite{popGCN}, InceptionGCN is proposed to use multi-size graph filters to improve the performance of GCN on disease prediction tasks.
\item{Multi-GCN~\cite{selfGCN}:} Multi-GCN applies the self-attention mechanism to aggregate the classification logits obtained from constructed multiple graphs corresponding to each of the demographic elements.
\item{LSTMGCN~\cite{LSTMGCN}:} Unlike~\cite{selfGCN}, LSTMGCN treats multi-modal representations obtained from multiple graphs as a sequence, and then aggregates them through an LSTM-based attention mechanism.
\item{EV-GCN~\cite{EV_GCN}:} The links of population graph in EV-GCN are computed through a learnable function of their non-imaging measurements.
\item{LGL~\cite{LGL}:} LGL tends to learn the graph based on the concatenated multi-modal features, and then perform classification based on the learned graph and multi-modal feature through GCN.

\end{itemize}

It should be noted that both PopGCN~\cite{popGCN} and InceptionGCN~\cite{InceptionGCN} are single-graph based methods and two of the earliest works to use GCNs for disease prediction tasks. Multi-GCN~\cite{selfGCN} is a multi-graph based method. In addition, we also compare with LGL~\cite{LGL} and EV-GCN~\cite{EV_GCN} which are the most related state-of-the-art works in disease prediction tasks.

\begin{figure*}[t]	
	\centering
	
	\includegraphics[width=4.5in]{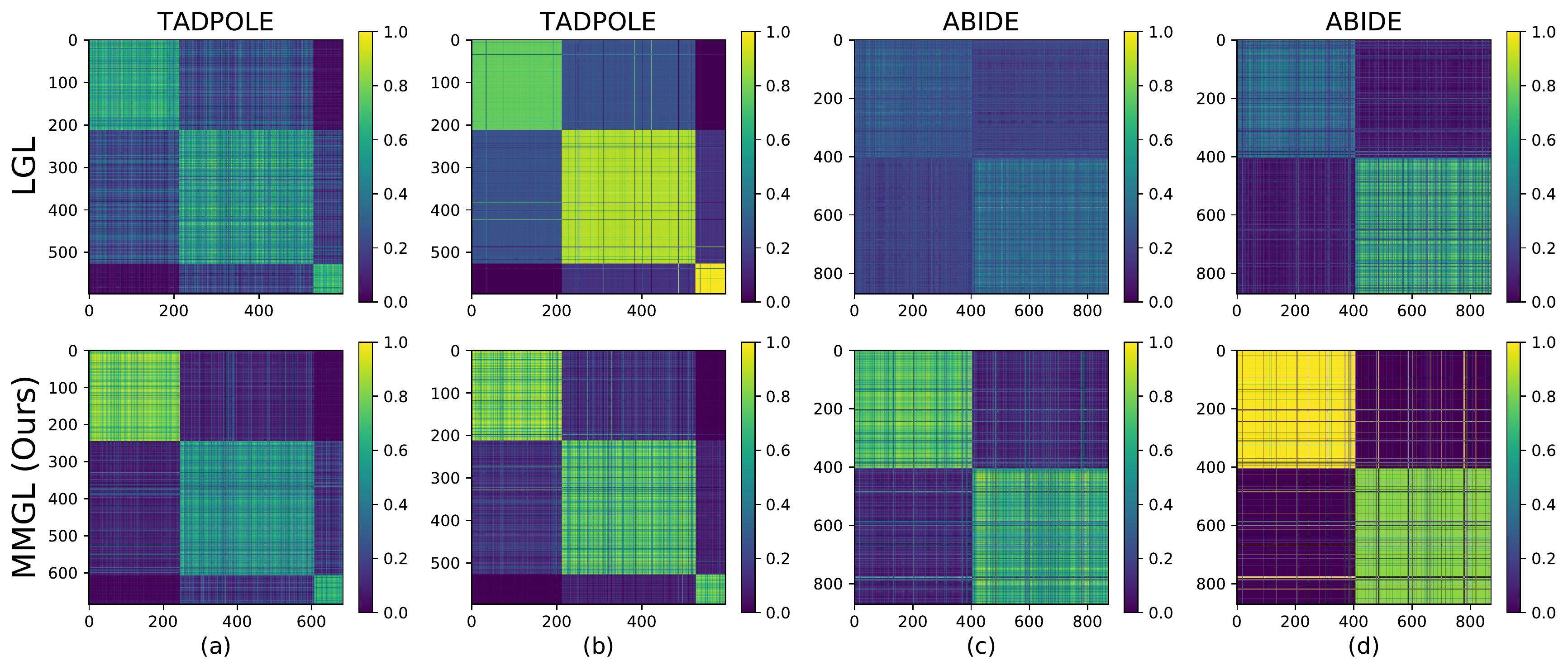}
	
	\caption{Visualization of the patient representation similarity matrices. (a) and (c) are the similarity matrices of \textbf{H} on two datasets, and (b) and (d) are the similarity matrices of aggregated representations through GNN on two datasets. The first row and the second row are the similarity matrices of the patient representation learned by LGL and MMGL, respectively.
	}
	\vspace{-0.5cm}
	\label{fig::contrast}
\end{figure*}

\begin{figure}[t]
	
	\subfigure[TADPOLE]{
		\centering
		\includegraphics[width=1.3in]{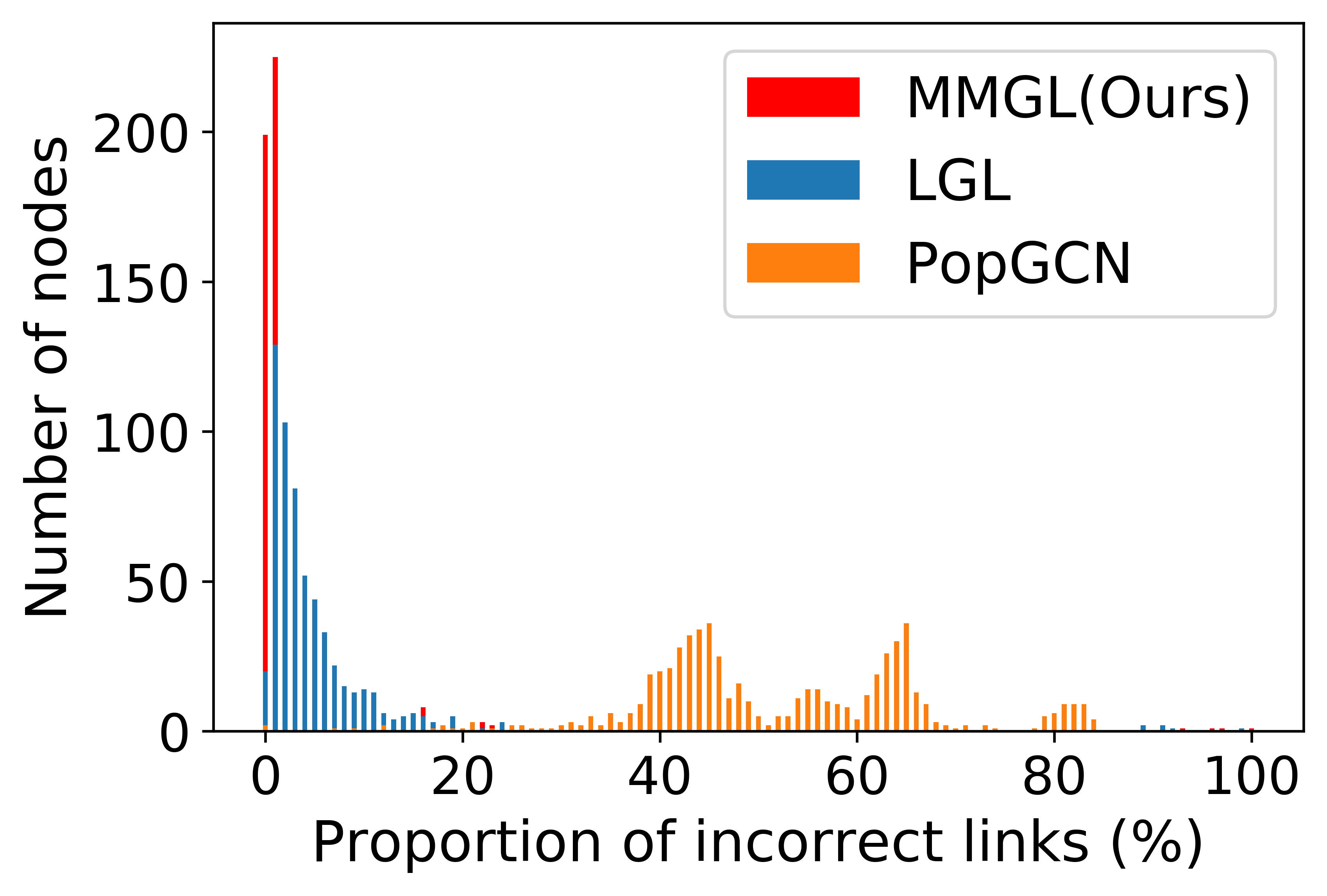}
		
	}%
	\subfigure[ABIDE]{
		\centering
		\includegraphics[width=1.3in]{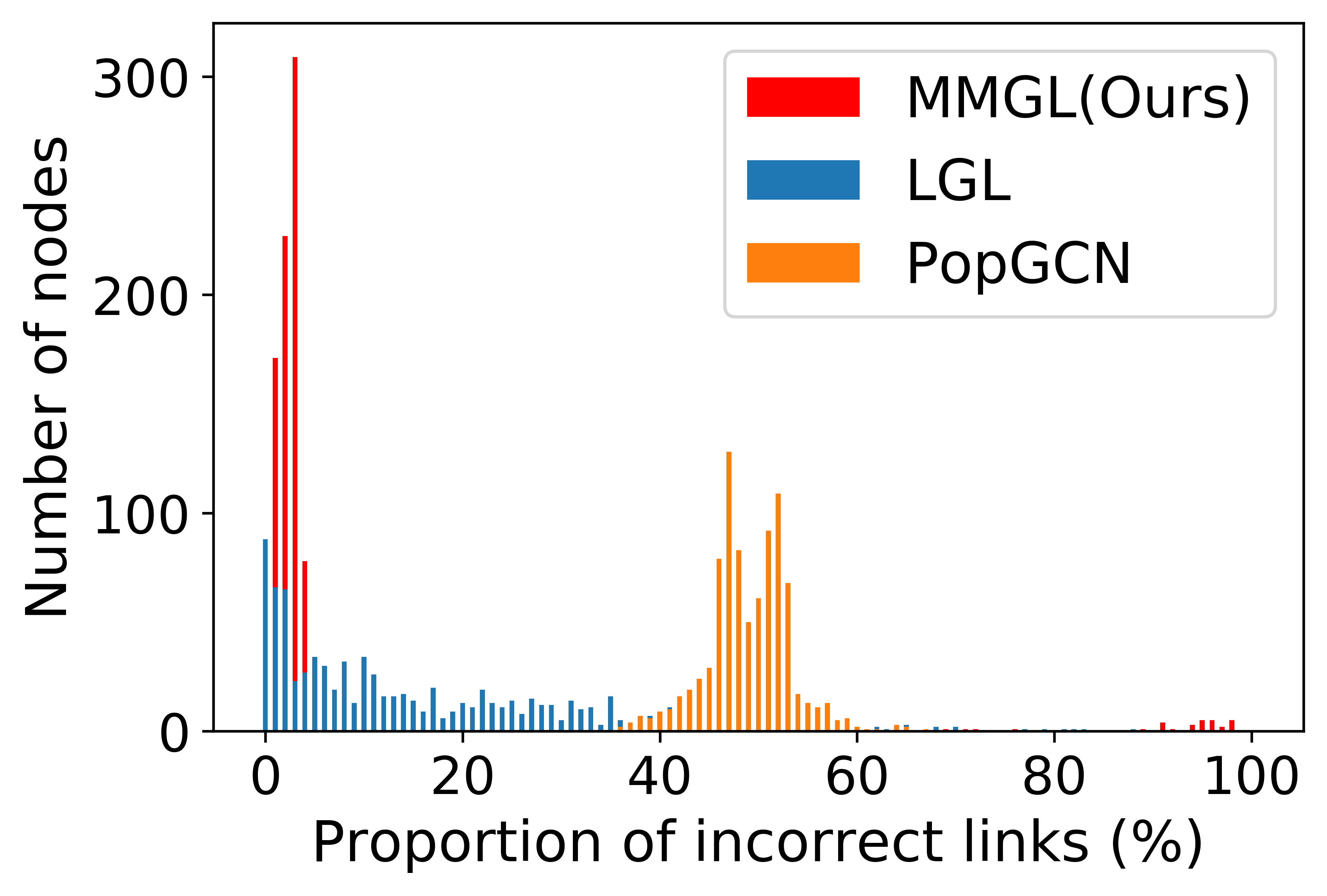}
		
	}%
	\centering
	\caption{Statistics of the incorrect link proportion for each patient in the population graph. For a patient in the graph, we calculate the percentage of incorrect links among all links of this patient.}
	\label{fig::link_proportion}
	\vspace{-0.5cm}
\end{figure}

\subsubsection{Quantitative Results}
We evaluate MMGL and the other compared methods on both the TADPOLE and ABIDE datasets using 10-fold stratified cross validation strategy. 
The mean scores, standard errors of Area Under Curve (AUC), and accuracy (ACC) are reported to show the performances of them. In addition, specificity (SPE) and sensitivity (SEN) are additionally provided to evaluate their performances on two-classification task. 

In Table~\ref{table::quantitative comparison}, the ``(H)" and ``(L)" prefix indicate that the method is based on handcrafted graph and graph learning, respectively. As shown in  Table~\ref{table::quantitative comparison}, we can conclude that,
\textbf{(i)} the strong contrast in the performance of handcrafted graph based methods on the two datasets points to the fact that such methods lack the ability of generalization for different datasets and tasks, which need much specific and tedious tuning to get a good performance. This drawback was also noted in ~\cite{InceptionGCN}. 
\textbf{(ii)} Compared to single-graph based methods (PopGCN, InceptionGCN) that simply use meta-features and imaging features, the further application of multi-modal features can effectively improve the performance of the model, such as LSTMGCN, EV-GCN. \textbf{(iii)} in three tasks, our MMGL significantly outperforms Multi-GCN on both TADPOLE and ABIDE datasets, which just verifies the effectiveness of MARL in capturing the inter-correlation among modalities. More importantly, the approximate $0.9\%$$\sim$$4.0\% $ improvement of our MMGL over LGL is achieved on ABIDE dataset. It means that AGL is more effective compared to the graph learning in LGL. 

Besides, to evaluate the performance on the transductive learning setting, we also report the performance of MMGL by increasing the number of patients labeled for training from 10 $\%$ to 80 $\%$ of the total number $N$ of patients, and the other patients are divided into two sets for validation and testing. It can be observed from Fig.~\ref{fig::multi-classificaion} that our MMGL performs the best compared to the baselines including PopGCN~\cite{popGCN} and LGL~\cite{LGL}. Meanwhile, compared with PopGCN, MMGL has better performance in the case of a small amount of training data, which indicates that the graphs learned by our AGL are more suitable for semi-supervised node classification than those manually constructed.
\footnotetext[1]{The reported results here are our re-implementation of the original algorithms since the source code is not available.}
\subsubsection{Qualitative Results} To give qualitative performance evaluation, we visualize the similarity matrix of the patient representations learned by LGL and MMGL respectively. For both datasets, the similarity matrices based on the learned representation have shown explicitly block effect as we can see from Fig.~\ref{fig::contrast}. Clearly, each block corresponds to different classes reflecting the differences between classes.
It can be seen that the representation \textbf{H} learned by MMGL is more distinctive compared to LGL. In other words, our MARL can indeed capture a more effective representation than direct concatenation adopted by LGL. Meanwhile, after aggregation through GNN, the intra-class similarity of representations is increased, demonstrating that a suitable graph does facilitate the performance of downstream tasks.

\subsection{Analysis on the Learned Graph}
To verify the quality of the learned graph, we consider the proportion of incorrectly predicted links for each patient in the population graph learned through MMGL and the other compared methods. Here, for a patient $u$, the predicted $e_{uv}$ will be taken as a mispredicted link if patient $v$ is not in the same class as patient $u$. 
Obviously, as shown in Fig~\ref{fig::link_proportion}, the graph obtained through MMGL performs very well with only a few incorrect links, and LGL~\cite{LGL} is slightly lower than that of MMGL, which verifies the superiority of the AGL adopted in MMGL to capture the latent relationships among patients. However, we can also notice that the proportion of incorrect links is relatively high for  PopGCN~\cite{popGCN} based on manually designed  graph structure. It indicates that simply constructing a graph based on demographic information may not be optimal. Besides, although a manually constructed graph may be consistent with human intuition,  it heavily dependents on the predefined similarity metric that needs to be carefully designed. On the contrary, the excellent performance of MMGL shows that it is feasible  to learn adaptively the graph structure in an end-to-end manner, and it has strong generalization ability.
\begin{table*}[t]
\centering
\scriptsize
\caption{ Quantitative evaluation of ablation studies on the TADPOLE and ABIDE datasets.}
\begin{tabular}{@{}l|cc|cccc@{}}
\toprule
\multirow{2}{*}{} & \multicolumn{2}{c|}{TADPOLE} & \multicolumn{4}{c}{ABIDE} \\ \cmidrule(l){2-7} 
& ACC                     & AUC                     & ACC                    & AUC                    & SEN                    & SPE  \\ \midrule
MMGL              
&\textbf{92.31$\pm$1.73}  &\textbf{93.91$\pm$2.10}  &\textbf{89.77$\pm$2.72} &\textbf{89.81$\pm$2.56} &\textbf{90.32$\pm$4.21} &\textbf{89.30$\pm$6.04}      \\ \midrule
MLP+AGL           
& 88.50$\pm$2.24              & 89.85$\pm$3.37             & 87.43$\pm$3.22      & 86.51$\pm$2.68     & 88.60$\pm$5.59     & 84.38$\pm$4.76     \\
Concat+AGL        
& 86.73$\pm$3.15              & 84.50$\pm$4.98             & 82.10$\pm$3.63      & 84.35$\pm$4.24     & 82.51$\pm$9.77     & 87.37$\pm$8.78     \\ \midrule
MARL       
& 89.12$\pm$3.31              & 90.84$\pm$2.63             & 86.48$\pm$2.99      & 86.40$\pm$3.30     & 85.34$\pm$3.57     & 87.45$\pm$3.14    \\
MARL+$G_{popGCN}$       
& 87.86$\pm$4.55              & 86.23$\pm$3.37             & 70.91$\pm$6.42      & 71.34$\pm$5.40     & 77.62$\pm$6.45     & 73.47$\pm$6.24     \\
MARL+$G_{kNN}$          
& 88.35$\pm$1.24              & 88.40$\pm$1.77             & 84.25$\pm$3.60      & 84.32$\pm$3.11     & 80.65$\pm$7.37     & 86.27$\pm$5.18    \\
\bottomrule
\end{tabular}
\label{table::ablation}
\vspace{-0.25cm}
\end{table*}

\vspace{-0.25cm}
\subsection{Visualization of the Modality-specified Representation}
Generally speaking, an effective representation learning method should preserve the original patient information well in the embedding space. To evaluate the capability of the modality-aware representation $\textbf{H}_{sp}$ more intuitively, we use t-SNE~\cite{tsne} to visualize it in a two-dimensional space for TADPOLE and ABIDE dataset, respectively.
As shown in Fig.~\ref{fig::contrast}, $\textbf{H}_{sp}$ can be clustered into several clusters corresponding to the classes in both datasets. In particular, for TADPOLE dataset, it can be observed that the order of the different stages of the disease is well maintained between clusters. Furthermore, for ABIDE dataset, the representation obtained through MMGL is much discriminative with small within-class scatter and large inter-class scatter. It means that $\textbf{H}_{sp}$ learned through the modality-aware representation learning precisely represents the intra-class similarity while capturing the inter-class differences.
\begin{figure}[t]
	
	\subfigure[TADPOLE]{
		\centering
		\includegraphics[width=1.3in]{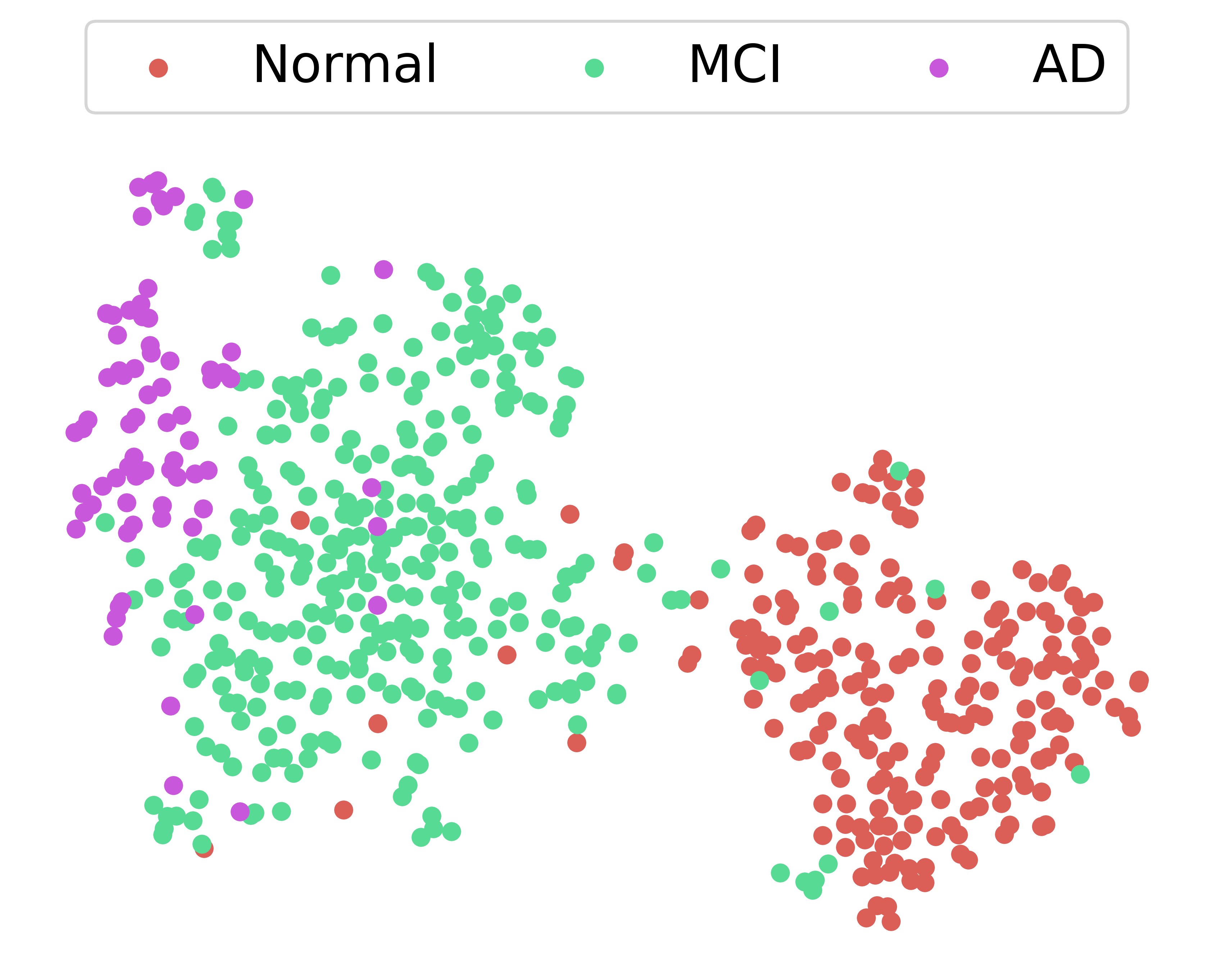}
	}%
	\subfigure[ABIDE]{
		\centering
		\includegraphics[width=1.3in]{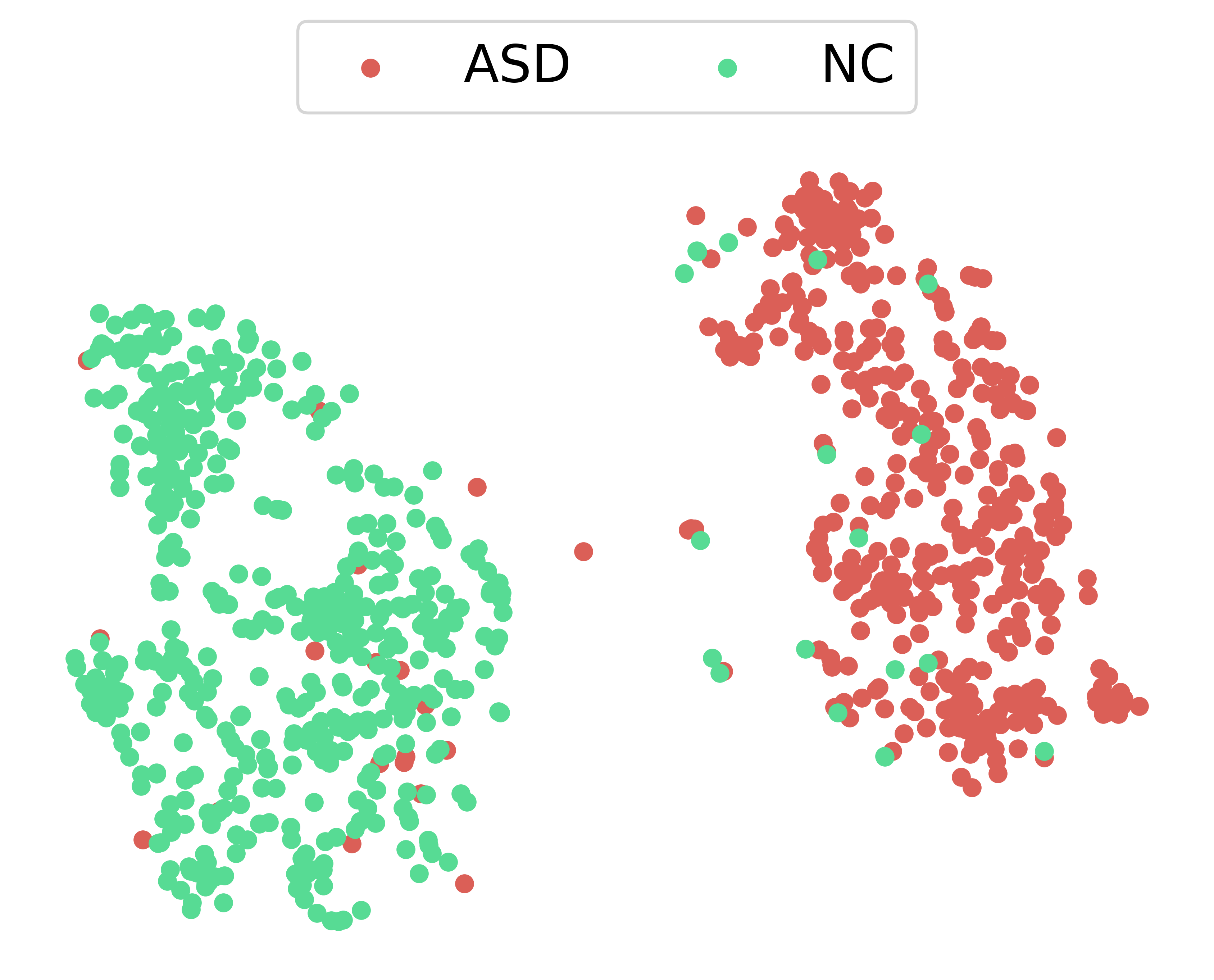}
	}%
	\centering
	\caption{Visualization of the modality-specified representation.}
	\vspace{-0.5cm}
\end{figure}

\begin{table}[]
\centering
\scriptsize
\caption{Ablation studies of AGL on transductive learning setting.}

\begin{tabular}{@{}c|cc|cccc@{}}
\toprule
\multirow{2}{*}{} & \multicolumn{2}{c|}{TADPOLE} & \multicolumn{4}{c}{ABIDE}             \\ \cmidrule(l){2-7} 
                  & ACC       & AUC      & ACC & AUC & SEN & SPE \\ \midrule
MARL +IPT         & 92.45         & 93.77        & \textbf{90.13}   & 88.97   & 88.06   & \textbf{90.68}   \\
MARL+AGL          & \textbf{92.80}          & \textbf{94.17}        & 90.12   & \textbf{91.88}   & \textbf{90.32}   & 89.30    \\ \bottomrule
\end{tabular}
\label{compare2IPT}
\end{table}

\begin{figure*}[t]
	\centering
	\includegraphics[width=5in]{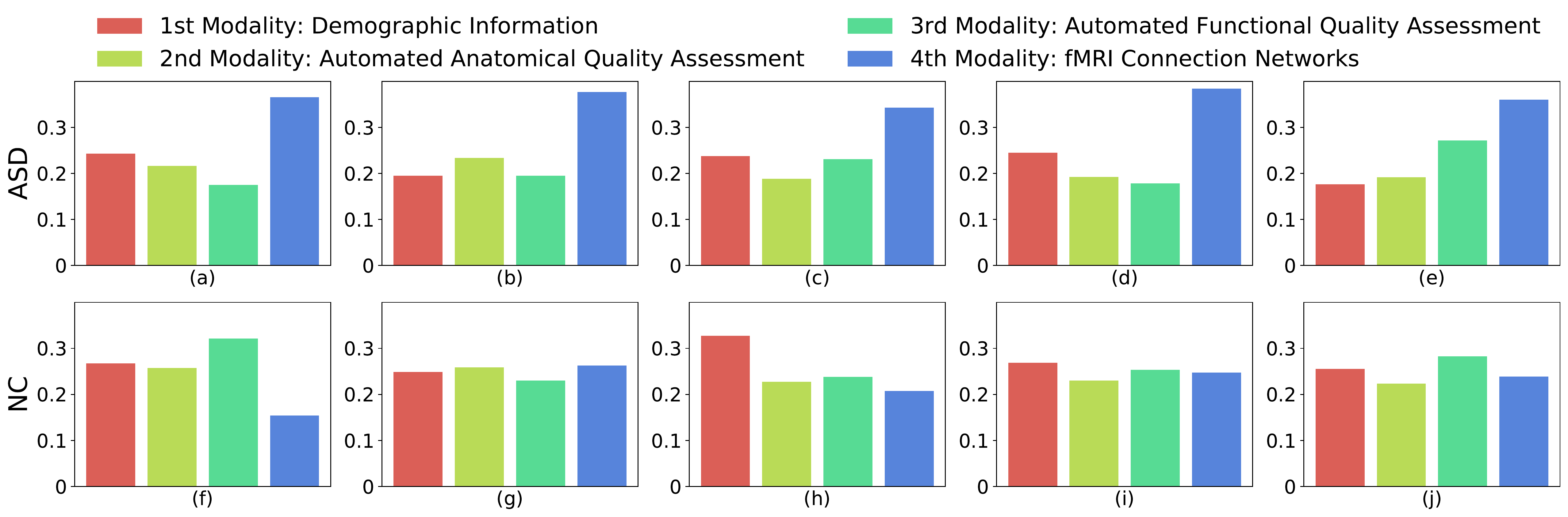}
	\vspace{-0.25cm}
	\caption{The contribution score of each modality obtained through MARL for ten subjects from ABIDE dataset. The vertical coordinate represents the contribution score. The first row and the second row are randomly selected from the ASD class and the NC class, respectively.
	}
	\vspace{-0.35cm}
	\label{fig::case}
\end{figure*}

\begin{figure}[t]	
	\centering
	
	\includegraphics[width=3in]{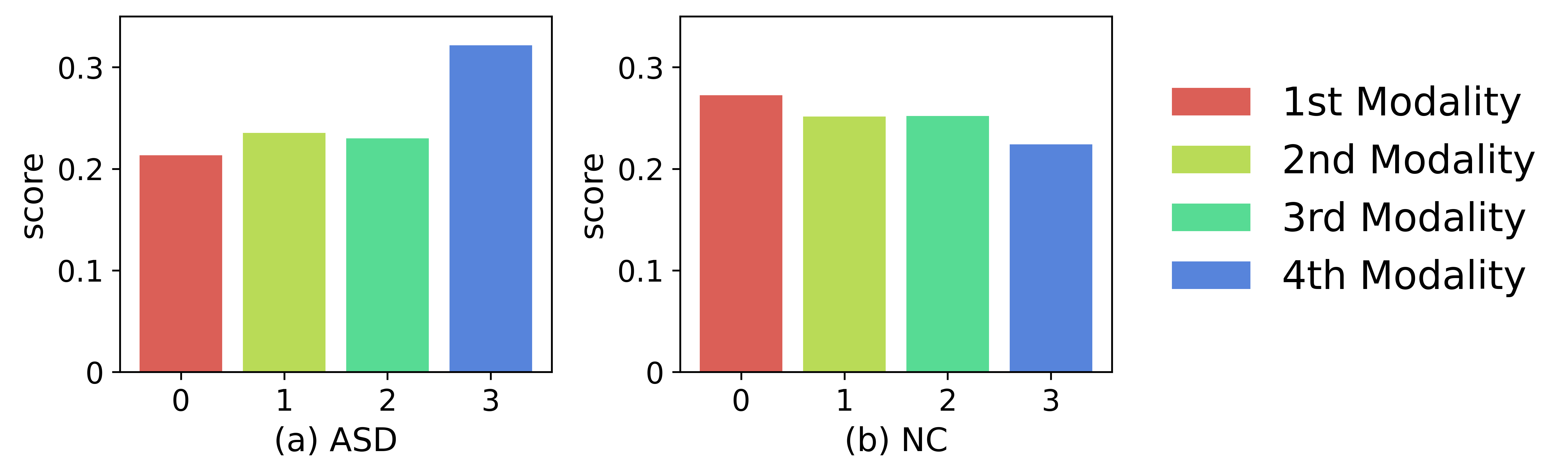}
	\vspace{-0.25cm}
	\caption{Visualization of the average contribution score of each modality for all patients from the NC class and the ASD class in ABIDE dataset, respectively.}
	\vspace{-0.5cm}
	\label{fig::case_avg}
\end{figure}
\vspace{-0.25cm}
\subsection{Ablation study}
To validate the effectiveness of modality-aware representation learning (MARL) and adaptive graph learning (AGL), we replace MARL with MLP and direct concatenation respectively, and replace AGL with the construction method adopted in popGCN~\cite{popGCN} and kNN graph $G_{kNN}$ using RBF kernel. Table~\ref{table::ablation} shows the ablation study results on applying different modules in our models. Specifically, the performance of the constructed graph $G_{PopGCN}$ of popGCN is the worst, especially on the ABIDE dataset, indicating that hand-constructed graph is indeed not a desirable choice. Furthermore, both ``MARL+$G_{popGCN}$'' and ``MARL+$G_{kNN}$'' worsen the performance of MARL, which means that an inappropriate metric can have negative impacts. Besides, AGL achieves a favorable performance despite the absence of MARL, which again validates the effectiveness of adaptive graph learning. More importantly, it can be observed that the combination of MARL and AGL achieves a performance that far exceeds the other combinations. In addition, we also compare the performance of AGL and transformer~\cite{transformer}, which is a popular architecture, as shown in Table~\ref{compare2IPT}. IPT is designed as an inter-patient transformer into which the patients’ representations are fed to obtain the results. It can be seen that the performance of the two methods is very close to each other.

\vspace{-0.25cm}
\subsection{Case Study}
To quantitatively describe the importance of each modality, we calculate the modality contribution scores for each patient based on the inter-modal attention matrix as it characterizes well the dependence between modalities. Specifically, for a patient $u$, the modality contribution scores $C_u$ could be obtained as $C_u = \frac{1}{M} \textbf{1}^{\top} \textbf{P}_{u}$, where $C_u \in \mathbb{R}^{1 \times M}$ and $C_{u,i}$ is the contribution score for the diagnosis of patient $u$ from the $i$-th modality.

We randomly select ten subjects from ABIDE dataset to give a case study by visualizing the contribution score of them. As illustrated in Fig.~\ref{fig::case}, it's obvious that the contribution score of the $4$-th modality (i.e., fMRI connectivity networks) is significantly higher than others for the subjects from the ASD class, indicating that fMRI connectivity networks play indeed a more significant role when the subject is diagnosed as ASD, which is quite consistent with the doctor's judgment in the actual diagnosis as pointed out in previous researches~\cite{fmri1, fmri2}.
In contrast, the distribution of the modality contribution scores of subjects from the NC class is relatively smooth and various, which may be due to the fact that the performance of the NC subjects are more dispersed than the ASD subjects.
Besides, we also present the average contribution scores for all subjects from the NC class and the ASD class, respectively. It can be observed from Fig.~\ref{fig::case_avg} that the average distribution of contribution score is roughly the same as that of the case distributions. It verifies that the modality-aware representation learning, especially the modality-specified representation induced from modality attention, could learn explainable representations, which may provide certain interpretations when making a diagnosis for the current patient.


\section{Conclusions}
In this paper, we propose a multi-modal graph learning framework named MMGL for disease prediction. To capture the shared and complementary information among multi-modality, we propose modal-aware representation learning to simultaneously obtain the modality-specified representation and the modality-shared representation considering inter-modal correlations. Furthermore, a lightweight adaptive graph learning is proposed to reveal the intrinsic relations among subjects, which could construct an optimal graph structure for downstream tasks. Meanwhile, MMGL could be jointly optimized in an end-to-end way, which enables more efficient training and inductive testing. Our ongoing research work will extend our MMGL to unified graph learning for incomplete data and more biomedical tasks.



\bibliographystyle{IEEEtran}
\bibliography{reference}

\end{document}